\RequirePackage{booktabs}
\documentclass[sn-mathphys-num]{sn-jnl}

\usepackage{graphicx}%
\usepackage{adjustbox}%
\usepackage{enumerate}%
\usepackage{multirow}%
\usepackage{amsmath,amssymb,amsfonts}%
\usepackage{amsthm}%
\usepackage{mathrsfs}%
\usepackage[title]{appendix}%
\usepackage{xcolor}%
\usepackage{textcomp}%
\usepackage{manyfoot}%
\usepackage{booktabs}%
\usepackage{algorithm}%
\usepackage{algorithmicx}%
\usepackage{algpseudocode}%
\usepackage{listings}%
\usepackage{algorithm}
\usepackage{algorithmicx}
\algrenewcommand\algorithmicrequire{\textbf{Input:}}
\algrenewcommand\algorithmicensure{\textbf{Output:}}
\usepackage{makecell}
\usepackage{tabu, booktabs}
\graphicspath{   {images/}
                }
\usepackage[caption=false]{subfig}
\usepackage{soul}

\theoremstyle{thmstyleone}%
\theoremstyle{thmstyletwo}%

\theoremstyle{thmstylethree}%
\DeclareMathOperator*{\argmax}{argmax}
\begin{document}

\title[Adversarial Attacks Guided by CLIP]{A Generative Adversarial Approach to Adversarial Attacks Guided by Contrastive Language-Image Pre-trained Model}


\author*[1,2]{\fnm{Sampriti} \sur{Soor}}\email{sampreetiworkid@@gmail.com}

\author[1]{\fnm{Alik} \sur{Pramanick}}\email{p.alik@iitg.ac.in}

\author[1]{\fnm{Jothiprakash} \sur{K}}\email{jothiprakash.k@alumni.iitg.ac.in}

\author[1]{\fnm{Arijit} \sur{Sur}}\email{arijit@iitg.ac.in}


\affil*[1]{\orgdiv{Department of Computer Science and Engineering}, \orgname{Indian Institute of Technology Guwahati}, \orgaddress{\postcode{781039}, \state{Assam}, \country{India}}}

\affil[2]{\orgdiv{School of Computer Engineering}, \orgname{Kalinga Institute of Industrial Technology Bhubaneswar}, \orgaddress{\postcode{751024}, \state{Odisha}, \country{India}}}


\abstract{The rapid growth of deep learning has brought about powerful models that can handle various tasks, like identifying images and understanding language. However, adversarial attacks, an unnoticed alteration, can deceive models, leading to inaccurate predictions. In this paper, a generative adversarial attack method is proposed that uses the CLIP model to create highly effective and visually imperceptible adversarial perturbations. The CLIP model's ability to align text and image representation helps incorporate natural language semantics with a guided loss to generate effective adversarial examples that look identical to the original inputs. This integration allows extensive scene manipulation, creating perturbations in multi-object environments specifically designed to deceive multilabel classifiers. Our approach integrates the concentrated perturbation strategy from Saliency-based Auto-Encoder (SSAE) with the dissimilar text embeddings similar to Generative Adversarial Multi-Object Scene Attacks (GAMA), resulting in perturbations that both deceive classification models and maintain high structural similarity to the original images. The model was tested on various tasks across diverse black-box victim models. The experimental results show that our method performs competitively, achieving comparable or superior results to existing techniques, while preserving greater visual fidelity.
}




\maketitle

\section{Introduction}
The advancement of deep learning in the past decade has led in its widespread application across various industries including self-driving cars, language translation, facial recognition, medical image analysis, satellite image analysis, speech recognition, improving image quality, and answering questions about images \cite{ghosh2025c,pramanick2025transdefensetransformerbaseddenoiseradversarial}.
However, in 2013, Szegedy et al. \cite{szegedy2013intriguing} discovered that neural networks' predictions can be manipulated by adding small noise-like perturbations, termed adversarial perturbations. These perturbations are typically invisible to the human eye but can drastically change the model's output. Understanding the effect of perturbations is crucial as it exposes vulnerabilities in machine learning models, particularly in safety-critical applications, and is essential for developing robust defense mechanisms to enhance the security and reliability of AI systems.

Adversarial attacks involve deliberate strategies designed to deceive machine learning models by introducing adversarial perturbations that cause incorrect predictions. Generally, for an image an adversarial perturbation $\delta$ is an additive noise vector applied to an input image  $x$ such that the perturbed image $x^\prime=x+\delta$ causes the image classification model $\mathcal{F}(.)$ to produce an incorrect output. Formally, adversarial attacks satisfy the condition $\mathcal{F}(x)\neq\mathcal{F}(x^\prime)$, where $\mathcal{F}(x)$ and $\mathcal{F}(x^\prime)$ represent the model’s prediction for the original image and the perturbed image respectively, subject to $||\delta||_p\leq\epsilon$, where $||.||_p$ is a norm (e.g., $\ell_2$ or $\ell_\infty$), and $\epsilon$ is a constant regulating the magnitude of the perturbation. The perturbation 
$\delta$ is often found by solving the following optimization.
\begin{equation}\label{eq2}
    \delta = \argmax_{\delta}{\mathcal{L}((x+\delta),y)}
\end{equation}
where $\mathcal{L}$ is the loss function used to train the model (e.g., cross-entropy loss) and $y$ is the true-label of $x$.

Adversarial attacks can be image-dependent and image-agnostic attacks. They can be executed by minimizing the perturbation value while ensuring successful deception of the model or by setting a maximum perturbation level and maximizing the model's fooling rate. 
The traditional methods for adversarial attacks can be broadly classified into two categories: gradient-based and non-gradient-based. Gradient-based methods utilize the gradient of the loss function to determine the optimal perturbation, while Non-gradient-based methods, on the other hand, use iterative methods, genetic algorithms, or optimization-based techniques, and do not rely on gradient information, which may be more complex to implement but can prove to be more effective against adversarial defenses.
\begin{itemize}
\item{\bf Gradient-based methods}
Fast Gradient Sign Method (FGSM) \cite{goodfellow2014explaining} attack uses the gradient information and pushes the input image in a negative gradient direction to misclassify it. This is a kind of white-box attack as it utilizes the model's information. FGSM is a major adversarial attack because it is computationally efficient. A lot of other extensions to these attacks have been proposed, The Fast Gradient Value Method (FGVM) of Rozsa et al. \cite{rozsa2016adversarial}, Iterative FGSM \cite{kurakin2018adversarial}, and Momentum Iterative FGSM \cite{dong2018boosting} etc.
Madry et al. proposed a Projected Gradient Descent (PGD) Attack \cite{madry2017towards}, which improves upon Iterative-FGSM \cite{goodfellow2014explaining}. Instead of clipping the perturbation, each iteration maintains imperceptibility by projecting perturbations onto an r-radius ball. It has effectively bypassed defenses like defensive distillation, making it a potent attack despite being computationally expensive. PGD-based attacks also address label leaking in adversarial training, a common issue with FGSM.

\item {\bf Non-Gradient Based Methods}
L-BFGS Attack \cite{szegedy2013intriguing} was the first to show deep learning model vulnerability to manipulation. Model deception is achieved by minimizing perturbation values. It establishes that inputs with slight differences in their L2 Norm distance metric are similar. The perturbation resembles the clean image, leading to incorrect model predictions. It employs a non-linear algorithm based on gradient values, solved through numerical optimization, which is effective in generating adversarial examples at the cost of significant computational resources.
The Carlini \& Wagner (C\&W) attack \cite{carlini2017towards} effectively bypassed the Defensive Distillation \cite{papernot2016distillation} defense. This uses low perturbation for fooling classifiers, making it resilient against defensive algorithms, increasing transferability across DL models, and ensuring the imperceptibility of perturbed examples. However, it suffers from high computational costs.
In \cite{papernot2016limitations} Nicolas Papernot et al. proposed a Jacobian-based Saliency Map Attack (JSMA) \& One-pixel attack, which iteratively adjust pixels until it is misclassified. A saliency map is crucial, computed from the Jacobian matrix to identify pixels influencing the target class likelihood. Similarly, the One-pixel attack \cite{su2019one} expands the adversarial importance map to alter selected pixels with maximum predictive impact.
The Deepfool Attack \cite{moosavi2016deepfool}, introduced by Moosavi-Dezfooli et al., minimizes perturbations by consistently nudging data points toward the decision boundary. 
Instead of generating separate perturbations for each input, the authors suggested Universal Adversarial Attacks \cite{moosavi2017universal}, which compute a single perturbation. When added to all the inputs, this perturbation causes the model to misclassify most of them.  Besides being image-agnostic, it exhibits transferability across models, making it truly universal. They demonstrated that a 4\% norm-bounded, quasi-imperceptible perturbation can achieve an 80\% fooling rate on popular ImageNet models (ResNet \cite{he2016deep}, Inception \cite{szegedy2015going}). Inspired by the Deepfool Attack, it involves repeatedly applying Deepfool to multiple images until the desired fooling rate is attained. 
\end{itemize}

Generative approaches to adversarial attacks are a relatively new and promising research direction that offer significant advantages over traditional methods. 
The Generative Adversarial Perturbations in \cite{poursaeed2018generative} (GAP) first popularized deep neural networks trained to maximize the likelihood of misclassification by generating such perturbed images. 
The Cross-Domain Transferability of Adversarial Perturbations as introduced in \cite{naseer2019cross} uses a domain-agnostic approach that reduces reliance on source data and launches highly transferable adversarial attacks. It uses relativistic loss to achieve scalability to large-scale datasets by learning a universal adversarial function and eliminating the need for expensive per-instance iterative optimization, which outperforms all existing attack methods by a significant margin, both instance-specific and agnostic. 
Recently, black-box domain transferability was explored by targeting low-level features of input images in Beyond Image-net Attack (BIA) \cite{zhang2022beyond}, where at first a random normalization module mimics diverse data distributions, boosting the attack's effectiveness regarding data, then, a domain-agnostic attention module captures essential features for perturbation, enhancing the attack's performance concerning models. The model handles images from any domain as input during inference, generating adversarial examples with a single forward propagation.

Existing adversarial attack methods are usually trained on single-object images due to the relative ease of generating adversarial examples compared to multi-object images. Methods using Gradient-based attacks are computationally expensive and have low transferability to different models or datasets. Additionally, non-gradient-based attacks are less effective than gradient-based ones and pose challenges in training due to their reliance on high-quality features for generating adversarial examples. Furthermore, real-world images frequently contain multiple objects within a scene, which contributes to the subpar performance of existing attack methods when applied to such images. Thus, this paper proposes a generative approach to adversarial attack guided by the CLIP model \cite{radford2021learning}, which is effective against both single-object and multi-object images, preserves high visual similarity between the original and perturbed images, and exhibits strong transferability across models.
%
%
\section{Background: Generative Adversarial Perturbations and Contrastive Loss }
The generative adversarial approaches use a generator model, typically an autoencoder, to create perturbations that are added to the original images to produce adversarial examples. The generator model is optimized using various loss functions to balance the need for visual similarity between the original and perturbed images with the goal of maximizing feature dissimilarity for successful misclassification. Features are extracted by one or more surrogate models, such as VGG \cite{simonyan2014very}, ResNet \cite{he2016deep}, or DenseNet \cite{huang2017densely} genre of models, which are similar to the target models (victim models in another term) in their ability to process and extract information from images. This training process is often sensitive to the hyperparameters of the models, the architecture of the generator model, and the correct feature extraction by the surrogate models among others. Despite these challenges, generative adversarial attacks have demonstrated improved effectiveness and hold great potential for advancing the field.

Current adversarial attack methods face two significant transferability challenges. Firstly, these methods frequently rely heavily on the availability and quality of the training data.
This dependence poses a substantial hurdle for "black-box" attacks, which aim to deceive models in diverse target domains without access to the target model's internal parameters or training data. 
Without direct training, crafting effective adversarial examples that generalize well across various models becomes exceedingly difficult, limiting the practicality of such attacks in real-world scenarios where training data is not accessible.
Secondly, while instance-agnostic attacks are computationally efficient and scalable, they generally exhibit weaker transferability compared to instance-specific approaches. Instance-agnostic attacks create a single perturbation pattern applicable to multiple inputs, which often results in lower success rates when transferred to different models. In contrast, instance-specific attacks generate perturbations tailored to individual inputs, leveraging the unique features of each input to maximize the likelihood of misclassification. Although more computationally intensive, instance-specific attacks typically offer better transferability across models because they exploit the intricacies of each input, leading to more robust adversarial examples. Balancing efficiency with effectiveness in transferability remains a key challenge in advancing adversarial attack methodologies.

A discriminator-free generative adversarial attack introduced as {\bf Symmetric Saliency-based Auto-Encoder (SSAE)} method in \cite{lu2021discriminator}, yielded high-quality results. 
Saliency maps are used to identify the most important features of the input data that significantly affect the model's predictions. By focusing on these salient features, the adversarial attack can be more efficient in terms of visual impercibility.
The SSAE method works by first using a saliency map to identify the critical regions of the input data. These regions are then manipulated by the generator in the GAN framework to produce adversarial examples. The auto-encoder component ensures that the generated adversarial examples are not only effective in fooling the target model but also remain visually similar to the original input, thereby maintaining their stealthiness.
SSAE's advantage lies in its ability to focus on crucial input features and be independent from a discriminator.

{The Contrastive Language Image Pre-trained (CLIP) Model} \cite{radford2021learning} by OpenAI bridges the gap between images and natural language.  Unlike traditional models specialized in either images or text, CLIP interprets and analyzes both simultaneously. It learns from a large dataset, pairing images with textual descriptions, enabling tasks like classification, object detection, and image generation. CLIP's architecture includes a vision encoder (based on CNNs) and a language encoder (based on transformers) that encodes input images and text into fixed-length vector representations, respectively. These two representations are then compared using cosine similarity. This similarity measure quantifies the alignment between the visual and textual features. A higher similarity score indicates a stronger correspondence between the image and text. The objective function used in CLIP is typically based on contrastive loss \cite{radford2021learning} like InfoNCE (Normalized Contrastive Estimation).

{\bf Contrastive loss} has proven to be highly effective in learning robust feature representations, leading to significant improvements in various applications like image retrieval, clustering, and anomaly detection. In {Contrastive Learning} \cite{oord2018representation} model learns meaningful representations by comparing and contrasting examples. This promotes similarity among matching image-text pairs that possess similar attributes or belong to the same class, whereas disentangling dissimilar ones. Training to differentiate between positive and negative pairs, helps the model to effectively capture and encode crucial task-related features. The contrastive loss is usually defined as a triplet loss function, which seeks to ensure that the distance between the anchor and positive sample is less than the distance between the anchor and negative sample by a certain margin. Given a pair of data $(x_i,x_j)$ and a binary label $y_{ij}$ indicating whether the points are similar ($y_{ij}=1$) or dissimilar ($y_{ij}=0$)
the contrastive loss can be defined as:
\begin{equation}
    \mathcal{L}_{contrastive} = \frac{1}{2} (\;y_{ij}).D^2 + (1-y_{ij}).max(\mu - D,0)^2\;)
\end{equation}
where $D=||f(x_i)-f(x_j)||_2$ is the Euclidean distance between the embedded representations 
$f(x_i)$ and $f(x_j)$ and $\mu$ is a predefined threshold that dictates how far apart dissimilar pairs should be.

GAMA \cite{aich2022gama} showcases the effectiveness of using the CLIP model to train powerful perturbation generators for scenes with multiple objects. By leveraging the combined image-text features, GAMA is capable of creating strong, transferable perturbations that deceive victim classifiers in diverse attack scenarios. Notably, GAMA induces around 16\% more misclassification compared to leading generative methods in black-box settings, where the attacker's classifier architecture and data distribution differ from those of the victim.
GAMA regulates the amount of perturbation only using the margin of contrastive loss, which experimentally shows that the perturbations spread across the image are often visually distinguishable as perturbed examples. In contrast, our method concentrates the perturbation over smaller regions of the image, making it less noticeable to the human eye.
%
%
\section{Contributions of this study}
In this study, we introduce a generative adversarial attack method that effectively combines the strengths of SSAE \cite{lu2021discriminator} and GAMA \cite{aich2022gama}. We conduct a thorough evaluation of the proposed method across different single and multi-object datasets, different pre-trained models, and different pretexts in CLIP to assess the robustness of the model's performance. Below are the highlights of the work presented here:
\begin{itemize}
    \item 
    {\bf Improved Visual Stealthiness:}
    We combine the concentrated perturbation approach from SSAE with the loss derived using the text and image feature-embeddings from CLIP, as utilized in GAMA, to generate adversarial examples that are both effective and less visually detectable.
    \item 
    {\bf Comprehensive evaluation across datasets and models:} 
    We validated our method on both single-object datasets (CIFAR-10 and Imagenette) and multi-object datasets (Pascal VOC), 
    and
    tested using different surrogate and target models, such as ResNet18 and DenseNet121, showcasing its robustness and adaptability across different model architectures. The effectiveness of our method was thoroughly compared against some state-of-the-art methods in terms of maintaining visual similarity and fooling the target models.
    \item {\bf Minimal Sensitivity to label Pretexts:} We compared different pre-texts along the class labels to generate CLIP text embeddings, showing that our method's performance is marginally affected by the choice of pretext, which underlines the robustness and reliability of our approach.
\end{itemize}
%
%
\section{Proposed method for adversarial attacks}
Our approach focuses on identifying the most vulnerable regions for attack by utilizing saliency maps, which highlight the areas of an image that are most influential for classification. Simultaneously, we ensure visual similarity between the original and perturbed images by minimizing the distance between their raw pixel values. To enhance the effectiveness of the attack, we further utilize contrastive learning by anchoring the features of the perturbed image to be dissimilar from those of the original image. This is achieved by leveraging CLIP’s text encoding, which is derived from randomly chosen labels, ensuring that the perturbed image's features deviate significantly from the raw image’s features. This dual approach not only targets the critical regions effectively but also maintains a high level of visual coherence, thereby increasing the likelihood of misleading the target classifier.
\begin{figure}[h!]
  \centering
    \includegraphics[width=\linewidth]{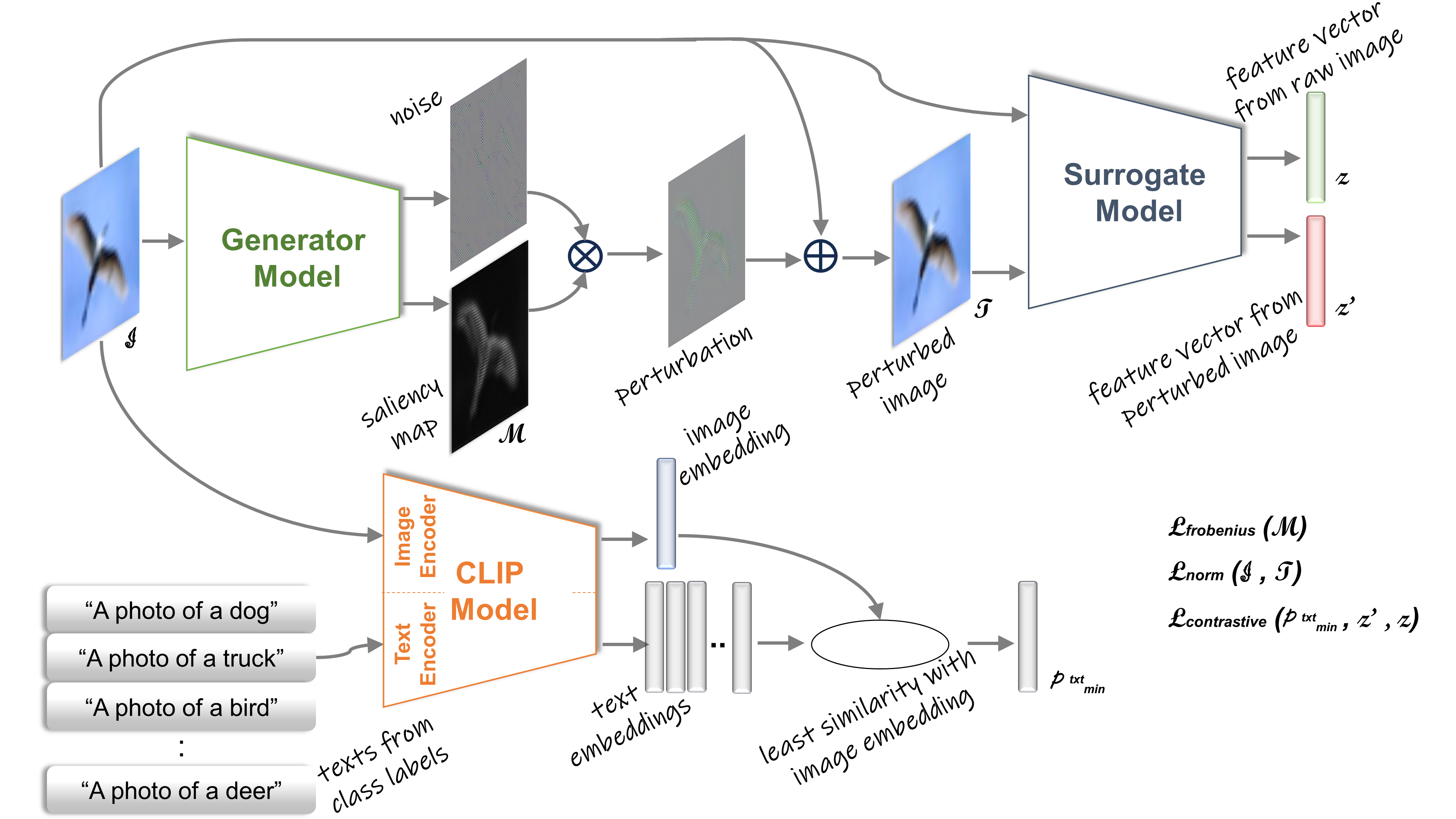}
    \caption{Architecture of the proposed adversarial attack model. Frobenius loss ensures concentrated perturbation, norm loss minimizes pixel-wise differences, and contrastive loss, with CLIP embeddings, ensures feature dissimilarity between raw and perturbed image. }
  \label{fig:fig1}
\end{figure}
\subsection{Proposed architecture}
The proposed architecture incorporates a generator $\mathcal{G}_{\theta}(\cdot)$, a pre-trained surrogate model $\mathcal{S}(\cdot)$, the text encoder of CLIP $T(\cdot)$, and the image encoder of CLIP $I(\cdot)$. 
The generator model $\mathcal{G}_{\theta}(\cdot)$ used in this study mirrors this architecture of the generator model detailed in the SSAE \cite{lu2021discriminator}, which features three key modules: an encoder, a perturbation decoder, and a saliency map decoder.
\begin{itemize}
    \item {\bf Encoder: }This component processes the input image ($\mathcal{I}$) through a sequence of layers: a 7×7 convolution, followed by two 3×3 convolutions, and six ResBlocks. This lightweight architecture effectively captures the essential features of the input.
    \item {\bf Perturbation Decoder:} It includes two 3×3 transposed convolutions and one 7×7 transposed convolution, generating perturbations ($\mathcal{P}$) of the same size as the input image. An additional constraint is applied to limit perturbation values to a maximum of 0.1, balancing attack effectiveness and visual quality.
    \item {\bf Saliency Map Decoder:} Similar in structure to the perturbation decoder but differing in its final layer, $\mathcal{G}_\theta$ produces a 1×W×H output ($\mathcal{M}$). This module helps identify critical regions for the attack by learning the relative importance of different areas, improving attack efficacy compared to using static masks. The saliency map approach avoids issues like visible attack boundaries and provides a more nuanced assessment of region importance.
\end{itemize}

The information flow in the proposed architecture, as shown in Figure \ref{fig:fig1}, is as follows.
\begin{enumerate}[i. ]
    \item A raw image $x$ is passed through the generator $\mathcal{G}_{\theta}(\cdot)$, resulting in a saliency-map $\mathcal{M}$ and a perturbation $\delta$. The perturbed image $(x + \delta)$ is then subjected to noise addition and projected into the same embedding space as the original image using the function $P(\cdot)$ from \eqref{eq2}, yielding the transformed image $x^\prime$.
    \item $I(\cdot)$ converts images into embeddings $\rho^{img} \in \mathbb{R}^K$, where $K$ represents the dimension of the embedding space. Image $x$ is passed through $I(\cdot)$ to obtain an image embedding $\rho^{img}$. 
    \item $T(\cdot)$ takes textual inputs and generates an embedding $\rho^{txt}_i \in \mathbb{R}^K$. From the set of possible labels, $M$ candidates $[{txt}_1, {txt}_2, ..., {txt}_M]$ are randomly sampled and are added with a fixed pre-text (e.g. "A photo of {label}") and are passed through $T(\cdot)$ to generate $[\rho^{txt}_1, \rho^{txt}_2, ..., \rho^{txt}_M]$. From this set, $\rho^{txt}_{min}$ is selected that minimizes the cosine similarity metric, $cs(\cdot)$, with the image embedding, computed as \eqref{eq3}:
    \begin{equation}
    \rho^{txt}_{min} = \min [cs(\rho^{txt}_1, \rho^{img}), cs(\rho^{txt}_2, \rho^{img}), ..., cs(\rho^{txt}_M, \rho^{img})]\label{eq3}
    \end{equation}
    \item Both the original image $x$ and the perturbed image $x^\prime$ are fed into a surrogate model $\mathcal{S}(\cdot)$ to obtain their respective latent representations $z$ and $z^\prime$, where $z, z^\prime \in \mathbb{R}^K$.
\end{enumerate}
Subsequently, the weights $\theta$ of the generator $\mathcal{G}_{\theta}(\cdot)$ are updated based on the computed loss functions using $\mathcal{M}$, $z$, $z^\prime$ and $\rho^{txt}_{min}$. The loss functions are described in detail in the next section (\ref{sec:loss_fn}).

\subsection{Loss Functions} \label{sec:loss_fn}
The objective is to reduce both visual dissimilarity and feature similarity between the raw and perturbed images. To achieve minimal visual dissimilarity, we concentrate the perturbation area as much as possible using {\it Frobenius loss} on the generated saliency-map, and reduce pixel-wise differences between the raw and perturbed images through {\it Norm loss}. For minimizing feature similarity, we employ {\it Contrastive loss} by anchoring the most dissimilar text embedding retrieved by CLIP. In this setup, the positive sample is the feature from the perturbed image, while the negative sample is the feature from the raw image.

\textbf{Frobenius Loss:} 
The Frobenius loss ensures that the perturbation is concentrated in the most salient regions, enhancing attack effectiveness. Before calculating the loss, the saliency map $\mathcal{M}$ is normalized to the range [0,1] using min-max scaling. The Frobenius norm of the scaled saliency map measures the extent of perturbation concentration.
\begin{equation}
    \mathcal{M}_{scaled}=\frac{\mathcal{M}--min(\mathcal{M})}{max(\mathcal{M})-min(\mathcal{M})}
\end{equation}
\begin{equation}
    \mathcal{L}_{frobenius} = \frac{1}{N} \sum\limits_{i=1}^N\;\;||\mathcal{M}_{scaled}||_2
\end{equation}
where $N$ is the number of images (in a batch).

\textbf{Norm Loss:} 
The Norm loss minimizes the pixel-wise differences between the norms of the features extracted by the surrogate model from raw ($\mathcal{I}$) and perturbed ($\mathcal{I}$) images to $z$ and $z^\prime$ respectively, aiming to reduce visual distortion. This ensures that the perturbations are subtle and less perceptible. The $\ell_2$ norm is used to quantify these pixel-wise differences.
\begin{equation}
    \mathcal{L}_{norm} = \frac{1}{N} \sum\limits_{i=1}^N\;\; \left| \;||z||_2-||z^\prime||_2\; \right|
\end{equation}

\textbf{Contrastive Loss:} Integrating transferable features into generated perturbations is vital to tackling training challenges with the generator $\mathcal{G}_{\theta}(\cdot)$ based solely on the fooling objective. Using CLIP-guided contrastive loss helps converge and promotes the creation of transferable perturbations. The Contrastive loss maintains feature similarity constraints by leveraging CLIP embeddings. It ensures that the feature vector of the perturbed image, $z^\prime$, is most similar to the most dissimilar text embedding, $\rho^{txt}_{min}$, compared to the feature vector of the raw image, $z$. This loss is critical for ensuring that perturbations are not only visually inconspicuous but also less detectable in feature space.
\begin{equation}
    \mathcal{L}_{contrastive} = \frac{1}{N} \;\;\sum\limits_{i=1}^N\;\; \left(\;||(z^\prime)-(\rho^{txt}_{min})||_2 + max(0,\mu-||(z^\prime)-(z)||_2)\; \right)
\end{equation}
where $(v)$ denotes the angle of a vector $v$ and $\mu>0$ is the maximum amount of perturbations.

\textbf{Total Loss:} ($\mathcal{L}_{\text{total}}$) is the sum of the fooling loss, text embedding guided loss, and image embedding loss, given as \eqref{eq7}.
\begin{equation}
\mathcal{L}_{total} = \alpha*\mathcal{L}_{frobenius} + \beta*\mathcal{L}_{norm} + \mathcal{L}_{contrastive}\label{eq7}    
\end{equation}
where $\alpha$ and $\beta$ regulate the learning rate of frobenius loss ($\mathcal{L}_{frobenius}$) and norm loss ($\mathcal{L}_{norm}$) respectively.
This optimizes generator parameters $\theta$ in training to produce perturbations that deceive the target model while possessing transferable features across various deep-learning tasks.

\section{Experimentation \& Results}
\subsection{Experimental Setup}
\subsubsection{Datasets}
In our experiments, we utilized both single-object and multi-object datasets to evaluate the effectiveness of our proposed generative adversarial attack method. For single-object classification tasks, we employed the CIFAR-10 and Imagenette datasets. The CIFAR-10 dataset contains 60,000 color images, each with a resolution of 32x32 pixels, distributed across 10 distinct classes, serving as a standard benchmark for image classification. Imagenette, a subset of the larger ImageNet dataset, includes approximately 13,000 images across 10 classes, providing a manageable dataset for rapid testing and evaluation. For multi-object tasks, we used the Pascal VOC dataset, renowned for its comprehensive annotations across 20 object categories, containing over 10,000 images, facilitating robust evaluation of object detection, segmentation, and classification methods. This combination of datasets allowed us to thoroughly assess the versatility and robustness of our approach across different types of visual data.

\subsubsection{Victim models}
Pre-trained classifiers were utilized as victim models to launch attacks by introducing adversarial perturbations. For the single-object datasets CIFAR-10 and Imagenette, we employed DenseNet121 and ResNet18 as both surrogate and target models (victim models). For the multi-object dataset Pascal VOC, the victim models included several state-of-the-art classifiers such as VGG16, VGG19, ResNet50, ResNet152, DenseNet121, and DenseNet169, with VGG19, ResNet152, and DenseNet169 serving as surrogate models. These models were selected because of their extensive use and demonstrated effectiveness in a wide range of computer vision tasks, allowing us to draw performance comparisons with previous works in the literature. Prior to the attacks, we analyzed the baseline performance of the victim models, measured by the Hamming Score (\%), to guide the subsequent adversarial attacks. The baseline performances along with results on perturbed images are presented in Table \ref{tab:tab1} for the single-object datasets and Table \ref{tab:tab2} for the multi-object dataset.


\subsubsection{Implementation Details}
The architectural framework of the generator network ($\mathcal{G}_{\theta}(\cdot)$ ) was adopted from prior established research \cite{lu2021discriminator}. The clamp limit $\mathcal{G}_{\theta}$, $\epsilon$ was set to 0.2, meaning all perturbations were constrained in $[-0.2,+0.2]$ range. 
The training configuration employed the ViT-B/16 model as the CLIP model.  
As the prerequisite of the  ViT-B/16 model, all images were resized to dimensions of 224 x 224 pixels. The ViT-B/16 model creates both text and image embeddings as 512-dimensional vectors. To ensure that the dimensions of $z$ and $z^\prime$ match, 512-dimensional features are extracted from the surrogate models. This is achieved by selecting a layer in the surrogate models that outputs a 512-dimensional feature on its first dimension. The remaining dimensions are averaged to produce 512-dimensional vectors from the surrogate models as well. This alignment ensures consistency in feature representation, facilitating effective computation of the discussed losses. The contrastive-loss margin ($\mu$) was set to 0.5. The learning rate of the frobenius-loss and norm-loss, $\alpha$ and $\beta$ were set to 0.00001 and 0.001 respectively. For the training of $\mathcal{G}_{\theta}$ AdamW optimizer \cite{adamw} was employed with a learning rate of 0.0001, and the batch size was set to 4. The results presented here were obtained after 50 epochs. The pseudo-code of the training process is given in Algorithm-\ref{algo}.
\begin{algorithm}
    \caption{Pseudo-code of the training process}
    \label{algo}
    \begin{algorithmic}[1]
        \Require dataset ($\mathcal{D}$), pre-texted class-labels ($C$), perturbation generator ($\mathcal{G}_\theta$) with clamp bound $\epsilon$, pretrained surrogate model ($\mathcal{S}$), pretrained CLIP-encoders for text ($\mathcal{C}_{text}$) and image ($\mathcal{C}_{image}$), contrastive-loss margin ($\mu$), epoch ($e$), batch size ($b$)
        \Ensure perturbation generator $\mathcal{G}_\theta$ with optimized weights $\theta$
        \State {Randomly initialize $\theta$}
        \State{Input $C$ to $\mathcal{C}_{text}$ and get $\rho^{txt}$}
        \Repeat $\;\;$for batch data $\mathcal{D}_b$($\subseteq\mathcal{D}$)\Comment{for each epoch for each batch}
            \State{Input raw image $x$ ($\in \mathcal{D}_b$) to $\mathcal{C}_{image}$ and get $\rho^{img}$}
    		\State{Compute $\rho^{txt}_{min}$ ($\in\rho^{txt}$) which is least similar to $\rho^{img}$}
    		\State{Input $x$ to $\mathcal{S}$ and compute mid-level embedding $z$}
    		\State{Input $x$ to $\mathcal{G}_\theta$ to generate perturbed image $x^\prime$}
    		\State{Input $x^\prime$ to $\mathcal{S}$ and compute mid-level embedding $z^\prime$}
    		\State {Compute loss $\mathcal{L}_{total}$}
    		\State {update $\theta$ using AdamW \cite{adamw} minimizing $\mathcal{L}_{total}$}
      \Until{ convergence}
    \end{algorithmic}
\end{algorithm}

\subsection{Results Analysis}

In multi-class classification, the Hamming score is a metric that assesses the performance of a classification model. It is computed as the proportion of correctly predicted labels relative to the total number of labels. In the context of adversarial attacks, a lower Hamming score indicates a more successful attack, as it reflects a higher degree of prediction inaccuracy. Thus, the objective is to minimize the Hamming score to demonstrate the effectiveness of the attack. Another important metric is the fooling rate, defined as the difference between the Hamming scores of the non-perturbed (raw) and perturbed datasets. A higher fooling rate signifies a more successful attack. Additionally, structural similarity (SSIM) is an important metric for assessing the visual quality of perturbed images. High SSIM values indicate that the adversarial images are nearly indistinguishable from the original images, ensuring that the perturbations are not only effective but also visually imperceptible.

\begin{figure}[!t]
  \centering
    \includegraphics[width=.95\linewidth,trim={0 .6cm 13cm 0},clip]{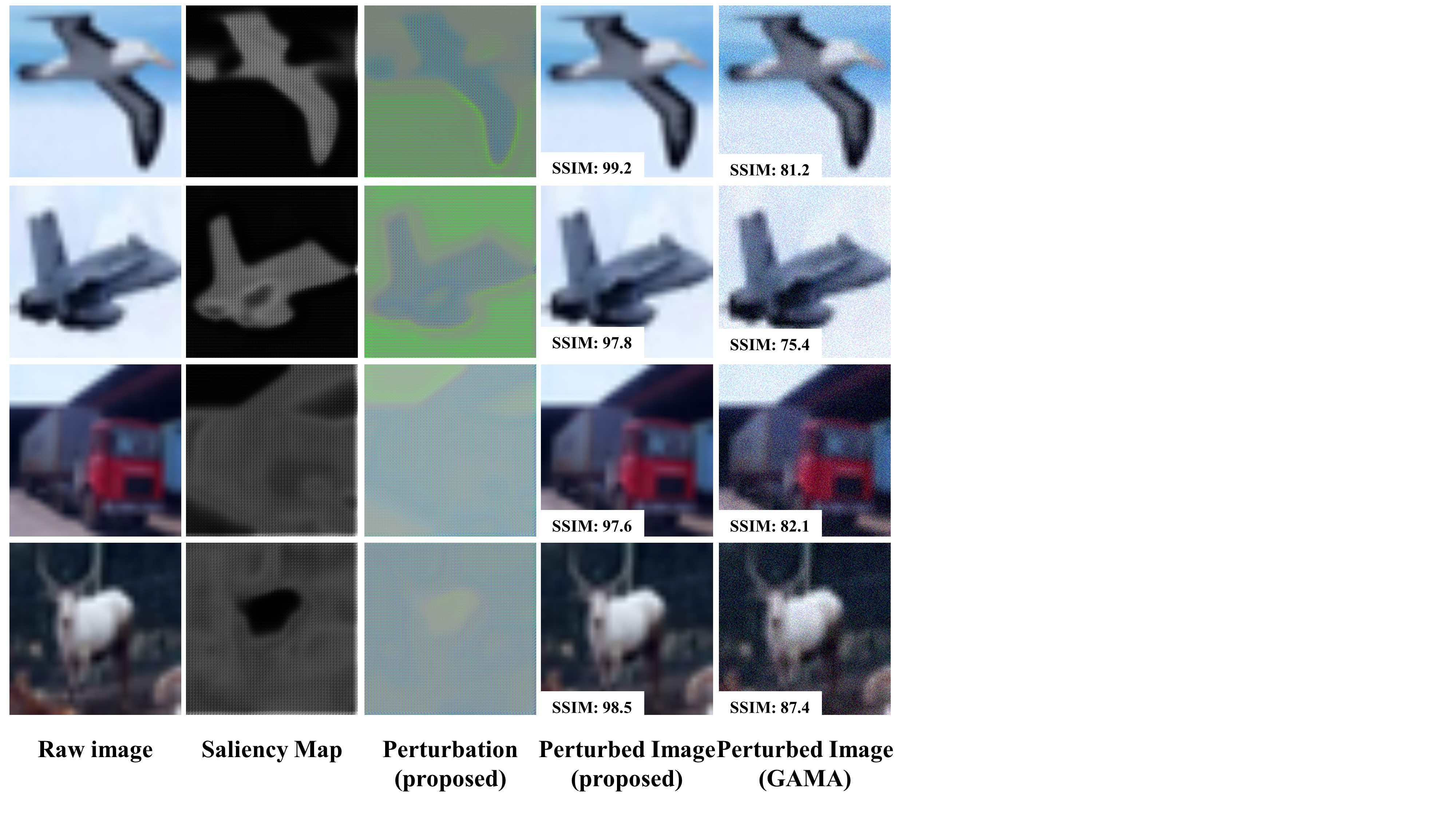}
    \caption{Examples from the CIFAR-10 dataset demonstrate that the proposed method maintains a high structural similarity between the perturbed images and the raw images, compared to the perturbed images generated by GAMA.}
  \label{fig:image_mosaic}
\end{figure}
\begin{table}[!thb]
  \centering
  \caption{Overall Hamming Score and Structural Similarity on CIFAR-10 and Imagenette Datasets by Different Adversarial Perturbation Methods Using Densenet121 and Resnet18 as both surrogate and target models}
  \label{tab:tab1}
  \begin{tabular}{cccccccc}
    \hline
\multirow{2}{*}{\makecell[c]{\bf Attacking\\\bf Methods}} &
&
\multicolumn{2}{c}{\textbf{Hamming Score (\%)} cifar10/imagenette } &
&
\multicolumn{2}{c}{\textbf{ssim} cifar10/imagenette} 
\\
\cline{3-4} \cline{6-7}
&&
\textbf{Densenet121} &  \textbf{Resnet18} &
&
\textbf{Densenet121} &  \textbf{Resnet18} 
\\
\hline\hline
No Attack&& 95.6/86.3 & 95.4/94.0 & & &  \\
\cline{1-1}\cline{3-4} \cline{6-7}
PGD&& 28.0/78.4 & 21.1/70.4 & & 1.00/1.00&  1.00/1.00\\
\cline{1-1}\cline{3-4} \cline{6-7}
FGSM&& 42.4/22.5 & 49.1/8.9 & & 0.96/0.88&  0.96/0.86\\
\cline{1-1}\cline{3-4} \cline{6-7}
SSAE&& 7.1/14.8 & 8.2/11.1 & & 0.97/0.97&  0.96/0.95\\
\cline{1-1}\cline{3-4} \cline{6-7}
GAMA&& 48.4/42.1 & 54.5/48.8 & & 0.74/0.71&  0.73/0.70\\
\cline{1-1}\cline{3-4} \cline{6-7}
Proposed&& 34.3/35.5 & 40.2/48.3 & & 0.92/0.96&  0.91/0.93\\
\hline
  \end{tabular}
\end{table} 

For our experiments, we tested the one-object datasets (CIFAR-10 and Imagenette) exclusively under white-box scenarios, where the surrogate and target models are the same. In contrast, the two-object dataset (Pascal VOC) was evaluated under both white-box and black-box scenarios, where the surrogate and target models differ. This dual approach enables a thorough comparison of the performance of our proposed method with that of previous models documented in the literature.

\begin{table}[!b]
\centering
\caption{Attack using different surrogate models on Pascal VOC dataset}
\label{tab:tab2}
\resizebox{\textwidth}{!}{
\begin{minipage}{\textwidth}
    \renewcommand*{\arraystretch}{1.33}
    \begin{tabular}{cc*{6}{c}}
    \hline
        \multirow{2}{*}{\makecell[c]{\bf Surrogate\\\bf model}} 
        &   
        \multirow{2}{*}{\rotatebox[origin=c]{00}{ \makecell[c]{\bf Attacking\\\bf Methods}}} 
        & \multicolumn{6}{c}{\textbf{Victim models (HS\%)}} \\
    \cline{3-8} 
        && 
        \textbf{VGG16} &  \textbf{VGG19} &  \textbf{ResNet50} &  \textbf{ResNet152} &  \textbf{DenseNet121} &  \textbf{DenseNet169}\\
        \hline
        \hline
        &No Attack&82.69&83.18&80.52&83.12&83.74&83.07\\
    \hline
        \multirow{5}{*}{VGG19} 
        &
        GAP & 19.64 & 16.60 & 72.95 & 76.24 & 68.79 & 66.50 \\
    \cline{2-8}
        &
        CDA & 26.16 & 20.52 & 61.40 & 65.67 & 70.33 & 62.67 \\
    \cline{2-8}
        &
        BIA & 12.53 & 14.00 & 64.24 & 69.07 & 69.44 & 64.7 \\
    \cline{2-8}
        &
        GAMA & 6.11 & 5.89 & 41.17 & 45.57 & 53.11 & 44.58 \\
    \cline{2-8}
        &
        Proposed & 5.85 & 5.51 & 45.73 & 50.40 & 57.36 & 51.05 \\
    \hline 
    \multirow{5}{*}{ResNet152} 
        &
        GAP & 56.93 & 56.20 & 65.58 & 72.26 & 75.22 & 69.54 \\
    \cline{2-8}
        &
        CDA & 41.07 & 47.60 & 53.84 & 47.22 & 67.50 & 59.65 \\
    \cline{2-8}
        &
        BIA & 45.34 & 49.74 & 51.98 & 50.27 & 67.75 & 61.05 \\
    \cline{2-8}
        &
        GAMA & 33.42 & 39.42 & 32.39 & 20.46 & 49.76 & 49.54 \\
    \cline{2-8}
        &
        Proposed & 19.61 & 27.73 & 31.65 & 27.03 & 38.63 & 50.29 \\
    \hline 
        \multirow{5}{*}{DenseNet169} 
        &
        GAP & 62.09 & 59.55 & 68.60 & 72.81 & 76.09 & 72.70 \\
    \cline{2-8}
        &
        CDA & 52.28 & 53.75 & 59.65 & 67.23 & 69.60 & 67.37\\
    \cline{2-8}
        &
        BIA & 48.52 & 53.77 & 56.15 & 63.33 & 54.01 & 58.85 \\
    \cline{2-8}
        &
        GAMA & 44.25 & 52.89 & 48.83 & 53.25 & 45.50 & 50.96 \\
    \cline{2-8}
        &
        Proposed & 48.85 & 55.89 & 54.39 & 59.25 & 51.49 & 58.01 \\
    \hline
    \end{tabular}
\end{minipage}
}
\end{table}


\begin{table}[!b]
\centering
\caption{Overall Hamming score (HS) and Fooling rate (FR) of the surrogate models used in Table-\ref{tab:tab2}}
\label{tab:tab5}
\resizebox{\linewidth}{!}{
\begin{minipage}{\textwidth}  
    \renewcommand*{\arraystretch}{1.33}
    \begin{tabular}{ccccccccccc}
    \hline
        \multirow{2}{*}{\makecell[c]{\bf Attacking\\\bf Methods}} &
        &
        \multicolumn{3}{c}{\textbf{Surrogate models (HS\%)}} &
        \multirow{2}{*}{\textbf{Avg}}&
        &
        \multicolumn{3}{c}{\textbf{Surrogate models (FR\%)}} &
        \multirow{2}{*}{\textbf{Avg}}
    \\
    \cline{3-5} \cline{8-10}
        &&
        \textbf{VGG19} &  \textbf{ResNet152} &  \textbf{Densenet169} &&
        &
        \textbf{VGG19} &  \textbf{ResNet152} &  \textbf{Densenet169} &
    \\
    \hline\hline
        GAP&& 
        53.45 & 65.95 & 68.64 & 62.68 &&
        29.27 & 16.77 & 14.08 & 20.03\\
    \cline{1-1}\cline{3-6} \cline{8-11}
        CDA&& 
        51.12 & 52.81 & 61.64 & 55.19 &&
        31.60&  329.90 & 21.07 & 27.53\\
    \cline{1-1}\cline{3-6} \cline{8-11}
        BIA&& 
        48.99  & 54.35 & 55.77 & 53.03 && 
        33.72 & 28.37 & 26.95 & 29.68\\
    \cline{1-1}\cline{3-6} \cline{8-11}
        GAMA&& 
        32.74 & 37.50  & 49.28 & 39.84 && 
        49.98 & 45.22 & 33.44 & 42.88\\
    \cline{1-1}\cline{3-6} \cline{8-11}
        Proposed&& 
        35.98 & 32.49  & 54.73 & 41.04 &&
        46.73 & 50.23 & 28.07 & 41.68\\
    \hline
    \end{tabular}
\end{minipage}
}
\end{table}




The results for the one-object datasets are compared with established adversarial attack methods such as FGSM \cite{goodfellow2014explaining}, PGD \cite{madry2017towards}, SSAE \cite{lu2021discriminator}, and GAMA \cite{aich2022gama} in Table-\ref{tab:tab1}. Our findings demonstrate that the proposed perturbation method, which utilizes the CLIP model, is as effective as or better than GAMA. Notably, our method maintains higher structural similarity than GAMA, making the perturbed images more indistinguishable from the human eye. This indicates that our approach not only successfully fools the classification models but also produces less detectable perturbations, enhancing the visual integrity of the adversarial examples. Figure 1 shows examples from the CIFAR-10 dataset, demonstrating that the proposed method maintains a high structural similarity between the perturbed images and the raw images, compared to the perturbed images generated by GAMA. Additionally, saliency maps and perturbations are presented for reference, highlighting the concentrated perturbation approach in minimizing visual disruption.

\begin{figure}[ht]
  \centering
    \includegraphics[width=.9\linewidth]{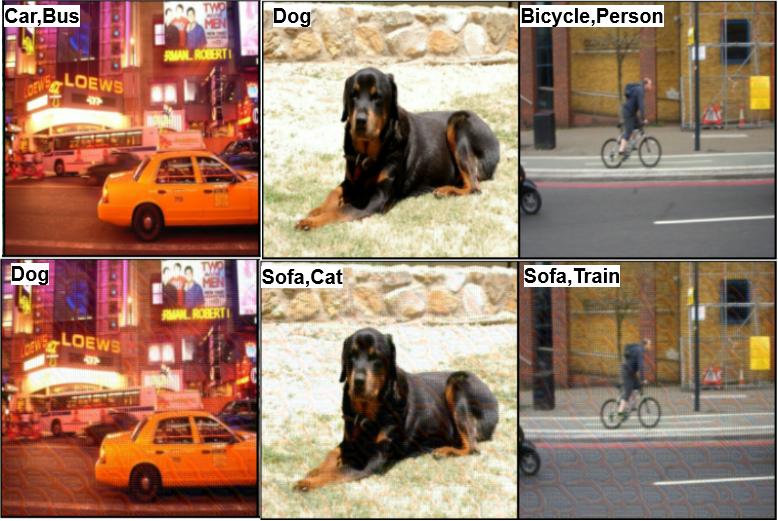}
    \caption{Qualitative examples illustrate a comparison between clean images (top row) and their corresponding perturbed images (bottom row) generated by the proposed method with samples taken from Pascal-VOC dataset. }
  \label{fig:fig3}
\end{figure}

The two-object dataset Pascal VOC is compared with GAP \cite{poursaeed2018generative}, CDA \cite{naseer2019cross}, BIA \cite{zhang2022beyond}, and GAMA \cite{aich2022gama}, where we noted the Hamming scores for different surrogate and target models, as detailed in Table-\ref{tab:tab2}. Our proposed method demonstrates better deceiving capabilities in almost all cases, highlighting its effectiveness across a variety of models and scenarios. 
In Table \ref{tab:tab5}, we summarize the results from Table \ref{tab:tab2}, highlighting the Overall Hamming Score (HS) and Fooling Rate (FR) of the surrogate models VGG19, ResNet152, and DenseNet169 for the Pascal VOC dataset. The fooling rates achieved by the proposed method are slightly lower than those of GAMA. However, it is experimentally observed that the perturbed images produced by our method maintain higher structural similarity with the raw images compared to those generated by GAMA. Additionally, with further fine-tuning and additional training epochs, the results could potentially be improved, enhancing both the fooling rates and the structural similarity of the adversarial examples. Figure \ref{fig:fig3} shows examples of misclassification in perturbed images from the Pascal-VOC dataset.
\begin{table}[!htbp]
    \centering
    \caption{Analysis on Text prompts for CLIP (dataset: Pascal-VOC, surrogate model: VGG19, target model: VGG16)}
    \label{tab:tab8}
  \renewcommand*{\arraystretch}{1.5}
    \begin{tabular}{cc}
        \hline
    \textbf{Text Prompt}&{\textbf{Hamming Score (\%)}} \\
\hline
\hline
\{Label 1\} and \{Label 2\}  & 6.35           \\
\hline
a \{Label 1\} and a \{Label 2\} & 6.32        \\
\hline
a photo depicts \{Label 1\} and \{Label 2\} & 6.11           \\
\hline
a photo of a \{Label 1\} and \{Label 2\} & 5.85          \\
\hline
    \end{tabular}
\end{table}



Additionally, a thorough analysis was conducted to evaluate 
the impact of various prefix prompts in combination with class names on the text encoder in CLIP was examined. The results of these experiments on Pascal-VOC dataset are presented in the following table \ref{tab:tab8}. The surrogate and target models were considered VGG19 and VGG16 respectively, for this experiment.
Despite varying the prompts alongside class names, CLIP's text encoder performance remained mostly consistent, indicating the limited influence of prompt choice on CLIP's text encoding capabilities on the final results.

\section{Conclusion and Future Work}
In this study, we proposed a generative adversarial attack method that uses the CLIP model to create imperceptible adversarial perturbations. By integrating the concentrated perturbation approach from SSAE with the dissimilar text embeddings similar to GAMA, our method effectively generates perturbations that are both impactful and visually indistinguishable from the original images. Our experiments demonstrated that the proposed method is as effective as, or surpasses, existing methods like GAMA in deceiving classification models. Specifically, our approach maintains higher structural similarity between the perturbed and raw images, making the adversarial examples less detectable to the human eye. We evaluated our method on several datasets, including CIFAR-10 and Imagenette for single-object classification, and Pascal VOC for multi-object classification. The results indicated that our method performs competitively across different models and datasets. The higher structural similarity of perturbed images compared to GAMA suggests that our approach offers a more refined perturbation that enhances visual stealthiness while maintaining effective adversarial impact.

Despite the promising results, there are numerous directions for future research to further improve the effectiveness and applicability of our method. First, the use of different vision transformers as surrogate models to potentially enhance the performance of adversarial attacks could be explored. For instance, experimenting with Vision Transformer (ViT) variants such as ViT-L/16 or ViT-H/14, which offer different embedding sizes and might capture richer features, could improve attack effectiveness. Additionally, investigating other CLIP backbones, like the OpenAI CLIP models with larger embedding dimensions, such as CLIP-RN50x64 or CLIP-RN101, could provide more powerful and discriminative embeddings for generating adversarial perturbations. Moreover, extending our approach to other computer vision tasks, such as object detection and segmentation, presents an exciting opportunity. Adapting our method to these domains will require modifications to handle complex outputs and could further demonstrate the versatility and robustness of the proposed technique. These advancements will be crucial for broadening the applicability and impact of generative adversarial attacks.
\bigskip
\bigskip
\backmatter
\bmhead{Supplementary information}



Not applicable
\bmhead{Acknowledgements}


This research work is supported by a Core Research Grant from
Science and Engineering Research Board (SERB), Department of Science and Technology (DST), Govt. of India vide DST No. CRG/2020/000651.
The authors would like to thank the associate editor and the anonymous reviewers for the constructive comments which improved the quality of this study.
\section*{Declarations}


\begin{itemize}
\item {\bf Conflict of interest/Competing interests:} 
The authors have no relevant financial interests in the manuscript and no other potential conflicts of interest to disclose.
\item {\bf Ethics approval and consent to participate:} Not Applicable
\item {\bf Consent for publication:} All the authors have read the final version of the manuscript and agreed for publication.
\item {\bf Data, Materials, and Code availability:} All the used data are cited at the intended places in the manuscript. 
\item {\bf Author contribution:} 
SS: Conceptualization, Methodology, Software, Writing - original draft; 
AP:  Conceptualization, Data curation, Writing - review \& editing; 
JK: Conceptualization, Methodology, Software, Investigation, Data
curation, Writing - original draft;
AS: Validation, Writing - Review \& Editing, Supervision.
\end{itemize}

\bibliography{sn-bibliography}

\end{document}